\documentclass[conference]{IEEEtran}
\IEEEoverridecommandlockouts
\usepackage{cite}
\usepackage{amsmath,amssymb,amsfonts}
\usepackage{algorithmic}
\usepackage{graphicx}
\usepackage{textcomp}
\usepackage{xcolor}
\usepackage{multirow}
\usepackage{pifont}
\usepackage{url}
\newcommand{\cmark}{\ding{51}}%

\def\BibTeX{{\rm B\kern-.05em{\sc i\kern-.025em b}\kern-.08em
    T\kern-.1667em\lower.7ex\hbox{E}\kern-.125emX}}
\begin{document}

\title{EasyRobust: A Comprehensive and Easy-to-use Toolkit for Robust and Generalized Vision
}

\author{\IEEEauthorblockN{1\textsuperscript{st} Xiaofeng Mao}
\IEEEauthorblockA{\textit{Alibaba Group}}
\and
\IEEEauthorblockN{2\textsuperscript{nd} Yuefeng Chen}
\IEEEauthorblockA{\textit{Alibaba Group}}
\and
\IEEEauthorblockN{3\textsuperscript{rd} Rong Zhang}
\IEEEauthorblockA{\textit{Alibaba Group}}
\and
\IEEEauthorblockN{4\textsuperscript{th} Hui Xue}
\IEEEauthorblockA{\textit{Alibaba Group}}
\and
\IEEEauthorblockN{5\textsuperscript{th} Zhao Li}
\IEEEauthorblockA{\textit{Zhejiang University}}
\and
\IEEEauthorblockN{6\textsuperscript{th} Hang Su}
\IEEEauthorblockA{\textit{Tsinghua University}}

}

\maketitle

\begin{abstract} Deep neural networks (DNNs) has shown great promise in computer vision tasks. However, machine vision achieved by DNNs cannot be as robust as human perception. Adversarial attacks and data distribution shifts have been known as two major scenarios which degrade machine performance and obstacle the wide deployment of machines ``in the wild''. In order to break these obstructions and facilitate the research of model robustness, we develop EasyRobust, a comprehensive and easy-to-use toolkit for training, evaluation and analysis of robust vision models. EasyRobust targets at two types of robustness: 1) Adversarial robustness enables the model to defense against malicious inputs crafted by worst-case perturbations, also known as adversarial examples; 2) Non-adversarial robustness enhances the model performance on natural test images with corruptions or distribution shifts. Thorough benchmarks on image classification enable EasyRobust to provide an accurate robustness evaluation on vision models. We wish our EasyRobust can help for training practically-robust models and promote academic and industrial progress in closing the gap between human and machine vision. Codes and models of EasyRobust have been open-sourced in \url{https://github.com/alibaba/easyrobust}. 

\end{abstract}


\section{Introduction}
Deep Neural Networks (DNNs) have achieved significant progress in various computer vision tasks, such as image classification~\cite{b1,b4} and object detection~\cite{b2,b3}. Despite their great success, there are concerns in the research community regarding the security of deep models. For example, DNNs are susceptible some imperceptible perturbations~\cite{b5} in input space, which are optimized by misleading the model to output wrong prediction. Such malicious inputs are also known as adversarial examples. In addition to adversarial examples crafted by humans, in real-world scenarios, deep models face more threats from test-time data distribution shifts under natural conditions~\cite{b6}. As we know, machine learning models including DNNs rely on a rigorous assumption that training and testing data are independent and identically distributed (i.i.d). However this ideal hypothesis is hardly satisfied in real world, where data distribution drift is frequently occurred due to image corruption~\cite{b7} or external environment changes, the model will have a significant decrease in performance and be exposed to potential security issues.

\begin{table}[htbp]
\caption{Open-sourced tools for robust computer vision}
\begin{center}
\begin{tabular}{|c|c|c|c|c|c|}
\hline
\multirow{2}{*}{\textbf{Tools}}&\multicolumn{2}{|c|}{\textbf{Benchmarks}} & \multicolumn{2}{|c|}{\textbf{Robust Method}}&\textbf{Analytical}\\
\cline{2-5} 
 & \textbf{\textit{Adv.}}& \textbf{\textit{OOD}}& \textbf{\textit{Adv.}}& \textbf{\textit{OOD}}&\textbf{Tools} \\
\hline
 CleverHans~\cite{b23} & \cmark &  & \cmark & & \\
 FoolBox~\cite{b24} & \cmark &  & \cmark & & \\
 ART~\cite{b25} & \cmark &  & \cmark & & \\
 Ares~\cite{b26} & \cmark &  & \cmark & & \\
 Robustness & \cmark &  & \cmark &  & \\
 RobustBench~\cite{b29} & \cmark & \cmark &  &  & \\
 model-vs-human~\cite{b30} &  & \cmark &  &  & \cmark\\
 DomainBed~\cite{b28} &  & \cmark &  & \cmark & \\
 RobustART~\cite{b27} & \cmark & \cmark & \cmark & \cmark & \\
 EasyRobust (Ours) & \cmark & \cmark & \cmark & \cmark & \cmark \\
\hline
\end{tabular}
\label{tab:tools}
\end{center}
\end{table}

\begin{figure*}[htbp]
\centerline{\includegraphics[width=0.9\linewidth]{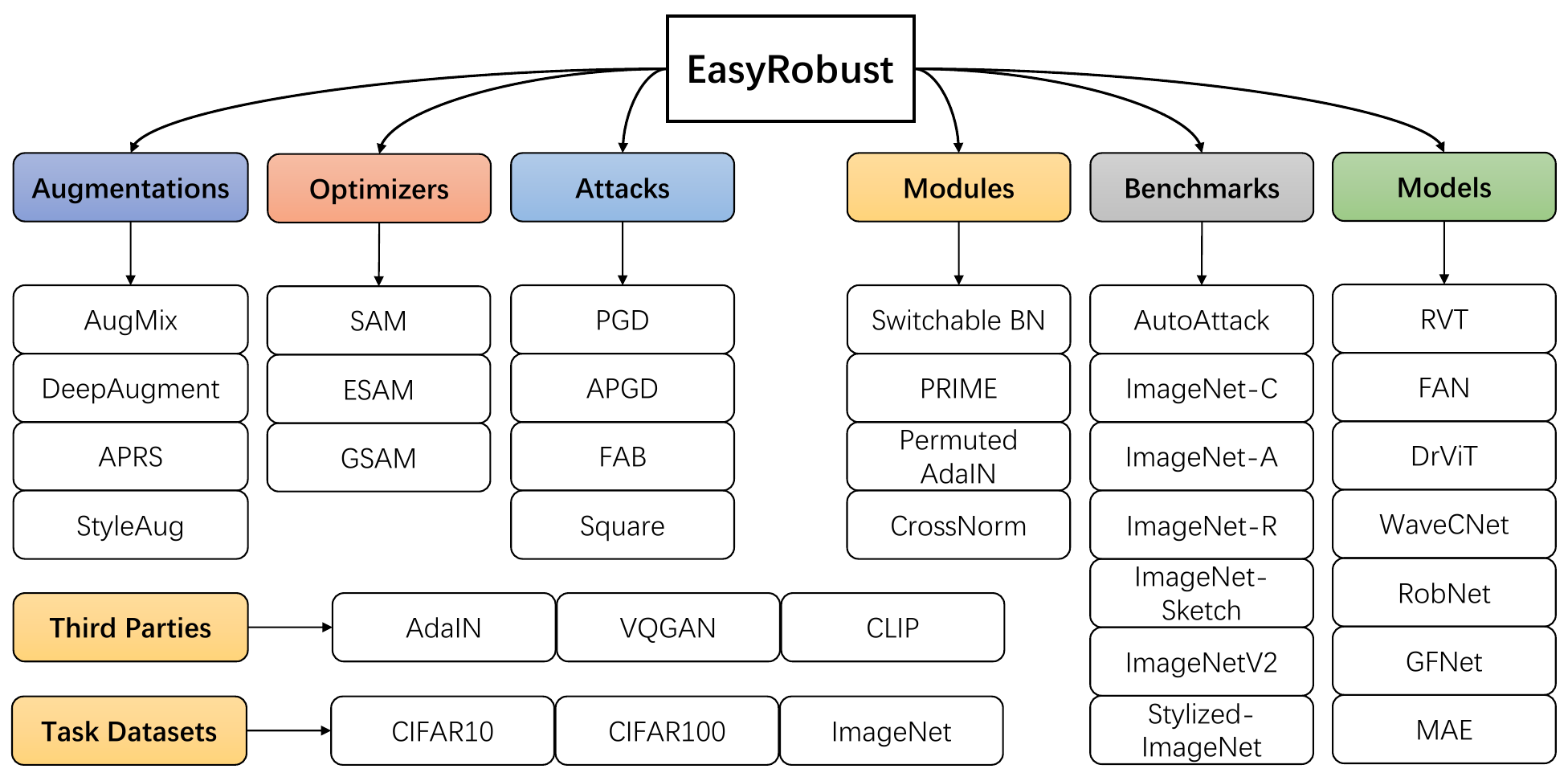}}
\caption{The overall structure of EasyRobust, which consists of six core components: augmentations, optimizers, attacks, modules, benchmarks and models.}
\label{fig:framework}
\end{figure*}

To overcome above problems, extensive benchmarks~\cite{b7,b8,b9,b10,b11,b12,b13} has been firstly established for accurate evaluation of model robustness. Promoted by these benchmarks, research community has proposed numerous methods to enhance the adversarial robustness or Out-Of-Distribution (OOD) robustness of DNNs. Specifically, Adversarial Training (AT)~\cite{b14} and preprocessing-based perturbation purification~\cite{b15} are two main ideas in defend of adversarial attacks. Based on AT, in recent year, stronger adversarial defenses are developed in ways of better objective~\cite{b16} and network architecture design~\cite{b17} and etc. Another branch of the OOD robustness methods use mix~\cite{b18} or style~\cite{b19} augmentation, AT-related techniques e.g. AdvProp~\cite{b20}, PyramidAT~\cite{b21}, DAT~\cite{b22} to improve the domain generalization ability. Most of them mitigate the traditional robustness and accuracy trade-off and yield enhancement on both In-Distribution (ID) and Out-Of-Distribution (OOD) performance.

There are also growing efforts for developing software packages to benchmark and build the robust models in computer vision. Early tools e.g. CleverHans~\cite{b23}, FoolBox~\cite{b24}, ART~\cite{b25}, Ares~\cite{b26}, Robustness\footnote{\url{https://github.com/MadryLab/robustness}} merely focus on the adversarial attacks and defenses on image classification models. With the degradation of performance under natural distribution shifts being gradually aroused attention, RobustART~\cite{b27} and DomainBed~\cite{b28} are proposed for better domain generalization. Some packages such as RobustBench~\cite{b29} and model-vs-human~\cite{b30} only support the benchmarking leaderboard and do not provide the implementation of methods. A comparison of existing robustness tools has been shown in Tab.~\ref{tab:tools}.

To accelerate research cycle in robust vision, we introduce EasyRobust, a Pytorch-based library for providing extensive tools studying the multi-dimensional model robustness. The major features of EasyRobust are: 1) \textbf{Comprehensive benchmark suite.} We integrate multiple test datasets and attack algorithms for robustness evaluation. 2) \textbf{Reproducible implementation of state-of-the-art methods in robust image classification.} 27 methods and more than 20 models have been open-sourced to accelerate further study based on existing robustness research. 3) \textbf{Scalability.} Large-scale ImageNet dataset is used as examplar task for training robust models. By supporting multi-node multi-gpu training and mixed precision training, EasyRobust can run on extra large tasks with huge training data. 4) \textbf{Analytical tools.} The analysis and visualization about the pre-trained robust models can further help us to explain how robust training improves the interpretability.

EasyRobust provides simple interfaces to users who want to evaluate the robustness of their models with just a few lines of code. A full set of standard hyper-parameters is used as the best practice for training ImageNet classification models. All supported methods are implemented based on these hyper-parameters, resulting in a clear and fair comparison. We also provide the robust pre-trained Swin-Transformer~\cite{b31} against adversarial attacks with 47.42\% accuracy under AutoAttack~\cite{b8}, and robust MAE-ViT-H~\cite{b32} against corruptions with 31.4 mCE on ImageNet-C dataset. These state-of-the-art models provide a strong foundation for researchers and algorithm engineers who are eager to apply pre-trained models in their robustness research. Through proper fine-tuning on our pre-trained models, the large training budget of robust model can be eased. For convenient usage, EasyRobust is designed to be dependent only on the existing popular libraries, including PyTorch, PyTorch Image Models (TIMM)\footnote{https://github.com/huggingface/pytorch-image-models}, both of which can be easily installed via PyPI. In this way, the advanced training techniques in TIMM can be compatible with our library and effect on the supported robust methods. A web-app hosted on ModelScope\footnote{https://www.modelscope.cn/home} makes the inference of EasyRobust models easier via executing directly from the browser.

The remaining sections are organized as follows. We first introduce the overall architecture of EasyRobust, various supported benchmarks and robust training algorithms and model visualization results by analysis tools. We then present the robustness benchmark results using our EasyRobust. Lastly, we discuss some potential limitations of our EasyRobust and possible expansion direction in the future.

\section{EasyRobust}
\subsection{Architecture}
\label{sec:2.1}
The core module of EasyRobust library contains six main components: attacks, augmentations, optimizers, modules, benchmarks and models. The applications, such as robustness training and evaluation are created on top of the basic functions of these six components. An overview is shown in Fig.~\ref{fig:framework}.
\subsubsection{Attacks} The attacks module includes basic algorithms used for AutoAttack~\cite{b8}, which benchmarks adversarial robustness of models. Specifically, AutoAttack consists of three white-box attacks (APGD-CE~\cite{b8}, APGD-DLR~\cite{b8}, FAB~\cite{b34}) and one black-box attack (Square)~\cite{b33}. Furthermore, our attack module also includes PGD~\cite{b14} attack. PGD is a classic attack algorithm used to evaluate the adversarial robustness, and it is often used in Adversarial Training (AT)~\cite{b14} to improve the model robustness.

\subsubsection{Augmentations} Training with diverse augmented images are common practice for avoiding over-fitting and improving generalization. For example, AugMix~\cite{b35} has figured out that through a mixing strategy of augmentation policies, the model can even be more robust under unknown image corruptions which have not seen during training. APR~\cite{b36} extends AugMix to frequency domain and uses augmented samples to force the model to pay more attention to phase components but keep robust to the variation of the amplitude. DeepAugment~\cite{b10} explores more diverse of augmentation via perturbing internal representations in an image-to-image network to create distorted images. Another type of augmentation is motivated by texture and shape feature. Deep model tends to use texture features while robust recognition needs more help from shape features. Thus StyleAug~\cite{b19} uses style transfer technique to create more augmented images with damaged texture feature and consistent shape feature, to make the model focus on shape features. We wish the supported data augmentation interface in EasyRobust can serve users for better handling of different OOD scenarios. 

\subsubsection{Optimizers} Deep model with a large number of parameters can easily fall into local minimum during optimization. Such phenomenon, also known as over-fitting, need to be prevented for better robustness and generalization. Recently, Sharpness Aware Minimization (SAM)~\cite{b37} is proposed to improve the traditional loss value based gradient descent optimizer with both minimizing loss value and loss sharpness. As a result, SAM performs efficiently for model generalization. Subsequently, numerous variants of SAM are proposed aiming at better efficiency and effectiveness. ESAM~\cite{b39} uses stochastic weight perturbation and sharpness-sensitive data selection strategies to accelerate SAM at no cost to its generalization performance. GSAM~\cite{b38} proposes a more precise measure of sharpness, named surrogate gap, which is easy to compute and consistently improves generalization. SAM and its variations are included in EasyRobust to assist users in training models avoiding local optima and attaining stronger generalization.

\subsubsection{Models} EasyRobust has integrated a variety of model architectures from CNNs to ViTs. Some of them are automatically discovered by Neural Architecture Search (NAS) algorithms. RobNet~\cite{b17} is the searched CNN model via optimizing its robustness against adversarial attacks. Instead of NAS, human priors can also be introduced as a kind of inductive bias that contributes to the robustness. WaveCNet~\cite{b41} replaces pooling layers of CNN model with Discrete Wavelet Transform (DWT) layer, thus to well keep object structures and suppress data noise during model inference. Recently, with the increasing usage of transformer structure in visual tasks, people begin to study the robustness of such structure and find that ViTs are inherently robust learners. RVT~\cite{b42} further extends the traditional ViT architecture with some empirical modification guided from the improvement of various robustness metric. Some robust ViT variants borrow the particularity of transformer structure. DrViT~\cite{b43} adopt a discrete design of visual token to make the token feature insensitive to changes in image details. FAN~\cite{b44} shows that self-attention promotes naturally formed clusters in mid-level representations via vision grouping, which is closely related to robustness under distribution shift. Thus it proposes to further strengthen vision grouping effect by incorporating attention mechanism into MLP layers.

\subsubsection{Benchmarks}
\label{sec:2.1.5}Based on different objectives, the benchmarks in EasyRobust can be divided into: 1) Adversarial robustness benchmarks and 2) OOD robustness benchmarks. The former uses the standard test set with perturbations crafted by adversarial attacks. AutoAttack~\cite{b8} is the most widely used attack method to evaluate the adversarial robustness. It consists of one white-box untargeted attack (APGD-CE)~\cite{b8}, two white-box targeted attacks (APGD-DLR~\cite{b8}, FAB~\cite{b34}) and one black-box attack (Square)~\cite{b33}. All attacks are ensembled by taking the worst case of model prediction. Since as one of the strongest attackers, AutoAttack has been accepted by researchers as an important standard for measuring adversarial robustness, our EasyRobust simply uses it as the standard metric and no longer implements other attack methods. For OOD robustness benchmarks, we adopt 7 test datasets based on ImageNet. Each of them represents a type of out-of-distribution scenario where the classifier is prone to make mistakes. ImageNet-A~\cite{b9} consists of natural images mis-classified by a ResNet-50. ImageNet-C~\cite{b7} applies a series of synthetic noise, blur, digital and weather corruptions on test images; ImageNet-R~\cite{b10} collects online images with artificial creation, e.g., cartoons, graphics, video game renditions, etc;
ImageNet-Sketch~\cite{b11} contains sketches images without detailed texture or color information. Stylized-ImageNet~\cite{b19} destroys the texture but maintains the shape feature by conducting style transfer on ImageNet images. ObjectNet~\cite{b12} places ImageNet objects in hard contexts, e.g. unusual backgrounds, rotations or imaging
viewpoints. Except for ImageNet-C which is measured by mCE, we report the top-1 accuracy on all above attacks and datasets. 

\subsubsection{Modules} EasyRobust contains some specially designed normalization or pooling layers for model generalization. 
\begin{table}[htbp]
\centering
\caption{Summary of robust methods in EasyRobust}
\begin{tabular}{| p{0.15\textwidth} | p{0.1\textwidth} | p{0.12\textwidth} |}
\hline
Category & Methods & Robust Scenarios \\
\hline\hline
\multirow{5}{*}{Augmentation-Based} & SIN & OOD    \\
                  &  APR         & OOD       \\
                  &  AugMix         & OOD       \\
                  &  DeepAugment         & OOD       \\
                  &  PRIME         & OOD       \\
                  \hline
\multirow{8}{*}{Training-Based}  & AT          & Adversarial      \\
                  &  AdvProp         &  OOD  \\
                  &  DAT         &   Adversarial, OOD    \\
                  &  HAT         &   Adversarial, OOD     \\
                  &  Debiased         &   OOD      \\  
                  &  WiSE-FT         &   OOD      \\
                  &  GNT         & OOD       \\
                  &  MAE         & OOD       \\\hline
\multirow{6}{*}{Model-Based}  & DrViT         &  OOD        \\
                  & RobNet         &  Adversarial        \\
                  &  FAN        &  OOD \\
                  &  WaveCNet         &   OOD      \\
                  &  RVT         &   Adversarial, OOD   \\
                  &  GFNet         &   OOD   \\\hline
\multirow{3}{*}{Optimization-Based}  & SAM         &  OOD        \\
                  &  GSAM        &  OOD \\
                  &  ESAM         &   OOD   \\\hline
\multirow{3}{*}{Normlization-Based}  & PermutedAdaIN         &  OOD        \\
                  &  SelfNorm        &  OOD \\
                  &  CrossNorm         &   OOD   \\\hline
\multirow{1}{*}{Pooling-Based}  & BlurPool         &  OOD      \\\hline
\multirow{1}{*}{Activation-Based}  & LP-ReLU         &  OOD      \\\hline
\end{tabular}
\end{table}
\label{tab:method}
Batch normalization is a frequently-used layer in CNNs. However, the mean and variance statistics of batchnorm will cause the model too dependent on the training data distribution, and limiting its generalization on OOD test data. To solve this problem, Permuted AdaIN~\cite{b45} and CrossNorm~\cite{b46} are proposed to introduce more variations on mean or variance statistics, via a way of style transfer among internal representations. Batchnorm also has potential weaknesses during adversarial training which will confuse the distribution statistics of clean and adversarial samples, resulting in the reduction of standard performance. Thus we develop a switchable batch normalization for users who want to conduct adversarial training but do not mean to affect the clean performance.

\subsection{Robust Methods in EasyRobust }
Tab.~\ref{tab:method} summarizes the robust methods supported in EasyRobust. Most of them  are implemented by pre-defined modules in core code of EasyRobust. Users can use these functions by simply importing corresponding modules into their private training code. For training-based methods, EasyRobust uses separate python scripts to show their complicated training process. Specifically, Adversarial Training (AT) enhances the adversarial robustness via a minimax optimization where inner step creates adversarial examples to maximize the loss, and outer step minimize the loss over generated adversarial examples. Some other works explore the effect of adversarial training in OOD generalization. 
\begin{table}[htbp]
\centering
\caption{Summary of robustness benchmarks in EasyRobust}
\begin{tabular}{| p{0.12\textwidth} | p{0.12\textwidth} | p{0.12\textwidth} |}
\hline
Category & Methods & Robust Scenarios \\
\hline\hline
\multirow{1}{*}{Attacks} & AutoAttack & Adversarial    \\\hline
\multirow{7}{*}{Datasets}  & ImageNet-A          & OOD      \\
                  &  ImageNetV2         &  OOD  \\
                  &  ImageNet-R         &   OOD    \\
                  &  ImageNet-Sktch         &   OOD     \\
                  &  ImageNet-C         &   OOD      \\  
                  &  Stylized-ImageNet         &   OOD      \\
                  &  ObjectNet         & OOD       \\\hline
\end{tabular}
\end{table}
\label{tab:benchmark}
AdvProp~\cite{b20} presents adversarial examples can be used to improve standard recognition accuracy with switchable batch normlization design. DAT~\cite{b22} conducts adversarial training in the discrete token space. With the rapid development of visual foundation models such as CLIP~\cite{b47}, people have paid more attention to the robust fine-tuning on downstream tasks. WiSE-FT~\cite{b48} is the first robust fine-tuning technique on pre-trained models, which finds the interpolation of weights between original and finetuned models can yield better performance. Except for training-based methods, other robust methods have been introduced in Sec.~\ref{sec:2.1}.

Up to the latest version, EasyRobust has implemented 27 robust methods. We will implement more methods and integrated them with EasyRobust in future development.

\subsection{Robustness Benchmarks in EasyRobust}
\label{sec:2.3}
Tab.~\ref{tab:benchmark} suggests the robustness benchmarks supported in EasyRobust. We have introduced them in Sec.~\ref{sec:2.1.5}. For AutoAttack, 5000 test images sampled from ImageNet validation set are used, which are provided by RobustBench. Adversarial perturbations with epsilon of 4/255 is applied on these images to get adversaral examples. Besides, there are also some benchmarks having not included in EasyRobust, such as DamageNet~\cite{b49} (a dataset consists of transfer-based black-box adversarial attacks), ImageNet-Patch~\cite{b50} (a dataset consists of adversarial patch attacks). We will integrate more useful benchmarks in future development of EasyRobust.

\subsection{Analytical Tools in EasyRobust}
EasyRobust supports several model analysis tools to help users to better explain and understand their robust models. A detailed illustration has shown in Fig.~\ref{fig:visualization}.
\subsubsection{Kernel Visualization}
\begin{figure}[htbp]
\centerline{\includegraphics[width=1.0\linewidth]{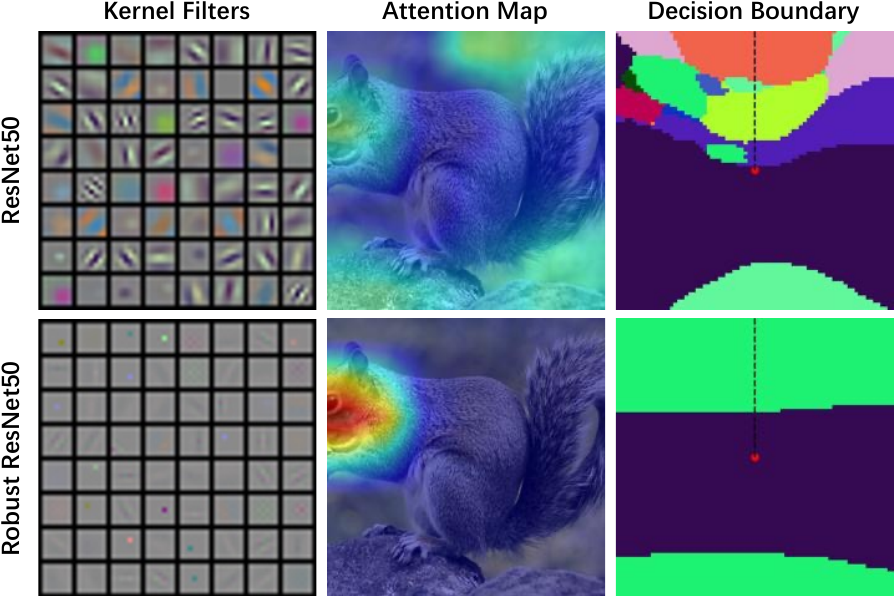}}
\caption{Examples of kernel filters, attention map and decision boundary visualization of ResNet50 models in EasyRobust. The normal ResNet50 uses the weights in torchvision, and the robust ResNet50 is obtained by adversarial training method.}
\label{fig:visualization}
\end{figure}

The convolution uses learned filters to convolve the feature maps from the previous layer. To visualize the first layer 7$\times$7 filters of a ResNet50~\cite{b1} trained on ImageNet-1K classification, EasyRobust can help users understand the model operation mechanism extracting features. In Fig.~\ref{fig:visualization}, standard CNN extracts edge, color and other low-level visual information in first layer. However, for a adversarially robust models, such detailed features are largely suppressed. It suggests that adversarially-robust models focus to more high-level features, such as image structure or object shape. For models with hundreds of kernel filters, EasyRobust supports a Principal Component Analysis (PCA) algorithm to reduce the number of filters visualized, and show the rough structure.

\subsubsection{Attention Visualization}
Attention map highlights the important region in the image that the model pays more attention to for getting final prediction. Class Activation Map (CAM)~\cite{b51} is a significant explainability method to get the attention of models. It uses internal activations to calculate which regions in the image are relevant to the target class. 
Following similar ideas, GradCAM~\cite{b52} and GradCAM++~\cite{b53} are using gradients or second order gradients for more flexible and accurate attention calculation. Our EasyRobust adopts grad-cam python library for attention map computing. A wide of attention visualization methods are supported, e.g., GradCAM~\cite{b52}, GradCAM++~\cite{b53}, XGradCAM~\cite{b54}, HiResCAM~\cite{b55}, ScoreCAM~\cite{b56}, AblationCAM~\cite{b57}, EigenCAM~\cite{b58}, EigenGradCAM~\cite{b58}, LayerCAM~\cite{b59}, FullGrad~\cite{b60} and etc. In addition, for models without spatial feature activations such as ViT, we also implement the attention rollout~\cite{b61} which is specialised for Transformers.

\subsubsection{Decision Boundary Visualization}
For a non-robust model, input test sample is more likely placed around the classification decision boundary, such that a small perturbation can lead the sample to go beyond the decision boundary and result in mis-classification. Thus it is meaningful that visualizing the class regions around the sample point and how far is the sample point from the nearest area with incorrect category. Our EasyRobust provides tools to realize this feature. We first obtain the adversarial gradient direction of the input sample as y-axis. The random direction orthogonal to the adversarial gradient is used as x-axis. By traversing all points in both directions, and showing the model predictions corresponding to these points, we can get the class region map in the right column of Fig.~\ref{fig:visualization}. We use different colors to present different class areas. Generally, a robust model should have less colors of class region and keep the largest area with the correct category. The visualization of our EasyRobust confirms the above general understanding. 

\section{Experiments}
\subsection{Settings}
\subsubsection{Datasets and metrics}
The benchmark experiments employ the datasets in Tab.~\ref{tab:benchmark}. Except for ImageNet-C which is measured by mCE, we report the top-1 accuracy on all used attacks and datasets.
\subsubsection{Basic training config for CNNs} 
\label{sec:3.1.2}The model is trained 90 epochs with a initial learning rate of 0.1 and gradually decayed with rate of 0.1 at 30, 60 epochs. SGD with momentum of 0.9 and weight decay of 0.0004 is adopted as the optimizer. The Nesterov momentum is enabled by default. The basic training augmentation is simple random resized crop and random horizontal flip. Input images are resize to 256$\times$256 and center cropped to 224$\times$224 at test time. Other tricks such as label smoothing, dropout are not used. The training is conducted on 8 GPUs, each of them has a throughput of 32 batch size of images.
\begin{table}
    \centering
    \caption{Benchmark Results of Adversarial Robust Methods in EasyRobust}
        \begin{tabular}{c|c|c|c|c}
        \hline
        Training & \multirow{2}{*}{Method} & \multirow{2}{*}{Model} & ImageNet & Auto \\
        
        Framework & & & Val & Attack \\
        \hline\hline
        EasyRobust & AT & Swin-B & 75.05 &	47.42 \\
        EasyRobust & AT & Swin-S & 73.41 &	46.76 \\
        Official & LightAT & XCiT-M12 & 74.04 &	45.24 \\
        EasyRobust & AT & ViT-B/16 & 70.64 &	43.04 \\
        Official & LightAT & XCiT-S12 & 72.34 &	41.78 \\
        EasyRobust & AT & EfficientNet-B3 & 67.65 &	41.72 \\
        EasyRobust & AT & ResNet101 & 69.51 &	41.04 \\
        EasyRobust & AT & ViT-S/16 & 66.43 &	39.20 \\
        EasyRobust & AT & EfficientNet-B2 & 64.75 &	38.54 \\
        EasyRobust & AT & ResNeSt50d & 70.03 &	38.52 \\
        Robustness & AT & WideResNet-50-2 & 68.46 &	38.14 \\
        EasyRobust & AT & ViT-B/32 & 65.58 &	37.38 \\
        EasyRobust & AT & EfficientNet-B1 & 63.99 &	37.20 \\
        EasyRobust & AT & SEResNet101 & 71.11 &	37.18 \\
        EasyRobust & AT & RX50\_32x4d & 67.39 &	36.42 \\
        Official & ViTvsCNN &  ViT-S/16 & 66.50 & 35.50 \\
        EasyRobust & AT & EfficientNet-B0 & 61.83 &	35.06 \\
        Robustness & AT & ResNet50 & 64.02 &	34.96 \\
        EasyRobust & AT & ResNet50 & 65.1 &	34.9 \\
        EasyRobust & AT & SEResNet50 & 66.68 &	33.56 \\
        EasyRobust & AT & DenseNet121 & 60.90 &	29.78 \\
        Official & Free-AT & ResNet50 & 59.96 &	28.58 \\
        Official & FGSM-AT & ResNet50 & 55.62 &	26.24 \\
        EasyRobust & AT & VGG16 & 59.96 &	25.92 \\
        EasyRobust & AT & VGG13 & 49.14 &	23.08 \\

        \hline
    \end{tabular}
    \vspace{-6mm}
    \label{tab:adv}
\end{table}
\subsubsection{Basic training config for ViTs}
\label{sec:3.1.3}The model is trained 300 epochs on 8 GPUs with total batch size of 1024. The learning rate first experiences 5 epochs warmup to 0.001, and then gradually decays to 0 with a cosine schedule. AdamW optimizer and 0.05 weight decay is adopted. We use RandAugment~\cite{b62}, ColorJitter, CutOut, CutMix~\cite{b63}, Mixup as training augmentation for ViTs. During inference, input images are resize to 256$\times$256 and center cropped to 224$\times$22

\begin{table*}
    \centering
    \caption{Benchmark Results of Non-Adversarial Robust Methods in EasyRobust}
        \begin{tabular}{c|c|c|c|c|c|c|c|c|c|c}
        \hline
        Training & \multirow{2}{*}{Method} & \multirow{2}{*}{Model} & ImageNet & ImageNet & ImageNet & ImageNet & ImageNet & ImageNet & Stylized & \multirow{2}{*}{ObjectNet} \\
        
        Framework & & & Val & V2 & C($\downarrow$) & R & A & Sketch & ImageNet &  \\
        \hline\hline
        EasyRobust & DAT & ViT-B/16 & 81.38 & 69.99 & 45.59 & 49.64 & 24.61 & 36.46 & 24.84 & 20.12 \\
        EasyRobust & - & RVT-S$^{*}$ & 82.10&	71.40&	48.22 & 47.84&	26.93&	35.34&	20.71&	23.24 \\
        Official & - & RVT-S$^{*}$ & 81.82&	71.05&	49.42&	47.33&	26.53&	34.22&	20.48&	23.11 \\
        EasyRobust & - & DrViT-S & 80.66&	69.62&	49.96&	43.68&	20.79&	31.13&	17.89&	20.50 \\
        Official & - & DrViT-S & 77.03&	64.49&	56.89&	39.02&	11.85&	28.78&	14.22&	26.49 \\
        Official & PRIME & ResNet50 & 76.91&	65.42&	57.49&	42.20&	2.21&	29.82&	13.94&	16.59 \\
        EasyRobust & PRIME & ResNet50 & 76.64&	64.37&	57.62&	41.95&	2.07&	29.63&	13.56&	16.28 \\
        EasyRobust & DeepAugment &	ResNet50 & 76.58&	64.77&	60.27&	42.80&	3.62&	29.65&	14.88&	16.88 \\
        Official & DeepAugment &	ResNet50 & 76.66&	65.24&	60.37&	42.17&	3.46&	29.50&	14.68&	17.13 \\
        EasyRobust & ANT & ResNet50 & 76.38&	63.99&	63.21&	39.51&	1.23&	26.75&	10.43&	16.57 \\
        Official & ANT & ResNet50 & 76.07&	63.85&	63.37&	39.02&	1.15&	26.35&	10.05&	16.54 \\
        EasyRobust & Augmix & ResNet50 & 77.81&	65.60&	64.14&	43.34&	4.04&	29.81&	12.33&	17.21 \\
        EasyRobust  & APR & ResNet50 & 76.28&	64.78&	64.89&	42.17&	4.18&	28.90&	13.03&	16.78 \\
        Official & Augmix & ResNet50 & 77.54&	65.42&	65.27&	41.04&	3.78&	28.48&	11.24&	17.54 \\
        Official  & APR & ResNet50 & 75.61&	64.24&	65.56&	41.35&	3.20&	28.37&	13.01&	16.61 \\
        Official & Debiased & ResNet50 & 76.91&	65.04&	67.55&	40.81&	3.50&	28.41&	17.40&	17.38 \\
        EasyRobust & SIN+IN & ResNet50 & 75.46&	63.50&	67.73&	42.34&	2.47&	31.39&	59.37&	16.17 \\
        Official & SIN+IN & ResNet50 & 74.59&	62.43&	69.32&	41.45&	1.95&	29.69&	57.38&	15.93 \\
        Official & AdvProp & ResNet50 & 77.04&	65.27&	70.81&	40.13&	3.45&	25.95&	10.01&	18.23 \\
        EasyRobust & Debiased & ResNet50 & 77.21&	65.10&	70.98&	38.59&	3.28&	26.09&	14.59&	16.99 \\
        EasyRobust & BlurPool & ResNet50 & 77.79&	64.81&	74.38&	37.56&	2.23&	25.98&	7.78&	18.28 \\
        Official & BlurPool & ResNet50 & 77.41&	64.73&	74.72&	37.00&	2.16&	25.10&	7.60&	18.14 \\
        EasyRobust & AdvProp & ResNet50 & 76.64&	64.35&	77.64&	37.43&	2.83&	24.71&	7.33&	16.82 \\
        EasyRobust & WiSE-FT & ViT-B/16 & 81.9 & 73.5 &	- &	79.1 &	52.6 &	53.6 &	- &	33.8 \\
        Official & WiSE-FT & ViT-B/16 & 81.7 &	72.7 & - &	78.8 &	52.2 &	53.2 &	- &	33.4 \\
        \hline
    \end{tabular}
    \vspace{-6mm}
    \label{tab:ood}
\end{table*}4. In addition, ViTs adopts label smoothing with 0.1, exponential moving average of weights and drop path for model regularization. Mixed precision training is turned on for ViTs.

\subsection{Benchmark Results for Adversarial Robustness}
Tab.~\ref{tab:adv} reports the benchmark results for adversarial robustness of the implemented methods in EasyRobust. Since the adversarial robustness benchmarks on CIFAR-10 and CIFAR-100 have been thoroughly presented in RobustBench. The benchmark results exhibited in our EasyRobust only focus on large-scale ImageNet classification task, which is more practical and general for industrial-scale applications. In experiments, attack epsilon is set as 4/255, clean performance is the top@1 accuracy (\%) on 50000 ImageNet validation images and robustness performance is the robust top@1 accuracy (\%) on 5000 AutoAttack adversarial examples based on sampled images provided by RobustBench. There is a shortage of pre-trained robust models on ImageNet task. Due to the expensive training cost, it is impractical to re-train the previous methods which have no open-sourced weights. Thus we collect 4 adversarially robust methods with weights open-sourced for comparison, namely FGSM-AT~\cite{b64}, Free AT~\cite{b65}, LightAT~\cite{b66} and ViTvsCNN~\cite{b67}. FGSM-AT and Free AT are fast adversarial training methods which meanwhile take the training time into consideration. In addition, we also compare the robust models trained by Robustness library. Various architectures including CNNs and ViTs are adopted for adversarial robustness study in EasyRobust. All adversarial training implementation of CNNs follow the config in Sec.~\ref{sec:3.1.2}. For ViTs, we slightly reduce the severity of data augmentation by deleting RandAugment and CutOut to prevent over-regularization. Previous works~\cite{b79} have demonstrated adversarial robustness will increase with the scaling size of model. Therefore, we do not use some extremely large models such as ViT-L~\cite{b40}, Swin-L~\cite{b31}, XCiT-L~\cite{b68}, for making a fairer comparison. 

From the results in Tab.~\ref{tab:adv}, the adversarially robust models trained by EasyRobust exhibit stronger performance. Compared with models with same size, a ViT-S/16 trained by our EasyRobust surpasses ViTvsCNN with +3.7\% on AutoAttack. On ResNet50, EasyRobust surpasses Free-AT and FGSM-AT with +5.14\%, +9.48\% on clean performance and +6.32\%, +8.66\% on adversarial robustness under AutoAttack. Compared with Robustness library, in the case of marginal gap on robustness (34.9\% vs. 34.96\%), EasyRobust achieves +1.08\% clean accuracy, exhibiting better robustness and accuracy trade-off. LightAT is based on a different network architecture (XCiT) which is not implemented in our EasyRobust. However, Swin-S has a similar scale with XCiT-M12. By comparing the robust Swin-S and XCiT-M12, we show our EasyRobust is slightly better than LightAT with +1.52\% on robust accuracy.

Through comparison of the robustness of different model structures, an apparently observation is that adversarially trained ViTs are showing priority with CNNs. The top@5 models in the ranking list are all belonging to ViT family. With similar model size, the ViT-S/16 obtains +1.33\% and +4.30\% higher than ResNet50 on clean and robust accuracy respectively. The worst model is VGG13~\cite{b71}, as an architecture proposed for early research, it only get 49.14\% clean accuracy and 23.08\% robust accuracy. By contrast, the best model Swin-B in EasyRobust achieves 75.05\% clean accuracy, which is even close to the performance of standard ResNet50 trained by clean samples. At the same time,  this model can maintain 47.42\% accuracy under AutoAttack. In addition, we also find that some tricks which is used to enhance standard performance of traditional CNNs may not contribute to robustness. A vivid exmaple is the Squeeze-and-Excitation (SE) module~\cite{b69}. Combining SE module into ResNet50 will result in a drop of 1.34\% robust accuracy under AutoAttack. Similar phenomena can also be found in ResNet101 model. However, as also an improvement of traditional CNNs, EfficientNet series~\cite{b70} are performed more effectively. EfficientNet-B3 gets both higher clean and robust accuracy than a ResNet50. Unfortunately, till now, there is no existing work to discuss in detail and rigorously about the robustness changes caused by the above network structural differences. Only some preliminary findings are revealed by existing works: 1) The non-smooth nature and abrupt change at 0 of ReLU will reduce the quality of the backward gradient, and do harms for adversarial training. Replacing ReLU with some smooth activations can alleviate this phenomenon~\cite{b72}; 2) Adversarial training can introduce difficulties of the batch normalization to model the different statistics of clean and adversarial examples. Such incompatibility will lead the models with batch normalization to underperform on adversarial robustness. A normalizer-free model can achieve adversarial robustness with only a minor sacrifice on clean sample accuracy~\cite{b73}.

\subsection{Benchmark Results for Non-Adversarial Robustness}
The non-adversarial robustness defines a set of OOD scenarios which occur naturally but not adversarially created. Tab.~\ref{tab:ood} shows the benchmark results for non-adversarial robustness of the implemented methods in EasyRobust. All the presented methods are sorted by corruption mCE on ImageNet-C with ascending order. We compare our EasyRobust implementation with 2 model-based methods: RVT-S$^{*}$~\cite{b42}, DrViT-S~\cite{b43}, 5 augmentation-based methods: SIN~\cite{b19}, APR~\cite{b36}, AugMix~\cite{b35}, DeepAugment~\cite{b10}, PRIME~\cite{b74}, 2 training-based methods: AdvProp~\cite{b20}, Debiased~\cite{b75} and 1 pooling-based method: BlurPool~\cite{b76}. For fair comparison, we modifies the training configs in Sec.~\ref{sec:3.1.2} and~\ref{sec:3.1.3} to keep consistent with their official implementation. The decay rates of exponential moving average are set as 0.99996 and 0.99992 for DrViT-S and RVT-S$^{*}$ respectively. APR and PRIME increase the training epochs to 100. AugMix adopt 180 training epochs and a cosine schedule for learn rate decay. DeepAugment expands the ImageNet train set two times with additional EDSR~\cite{b77} and CAE~\cite{b78} distortion set. So its number of epochs is decreased to 30 to keep overall optimization steps invariant. AdvProp turns off the Nesterov momentum in SGD and uses 105 epochs for training. Debiased warms up the learning rate at first 5 epochs, the overall training epochs are set as 100. WiSE-FT~\cite{b48} is a robust fine-tuning method of pre-trained models. Thus it only requires 10 training epochs with learning rate of 0.00003 and weight decay of 0.1. WiSE-FT additionally clips the gradient norm with 1.0. The best interpolated model is used for reporting the results in Tab.~\ref{tab:ood}. WiSE-FT does not evaluate on synthetic OOD datasets such as ImageNet-C and Stylized-ImageNet in official implementation, so these results are not reported in Tab.~\ref{tab:ood}. 

The results suggest that our EasyRobust can achieve better generalization ability on most compared methods. Particularly, EasyRobust implementation on DrViT-S can surpass the official version with +3.63\% on clean accuracy, and achieve superior results on all other OOD test datasets except for ObjectNet. For other methods such as RVT-S$^{*}$, DeepAugment, ANT, AugMix, APR, SIN, BlurPool and WiSE-FT, our EasyRobust can successfully reproduce their results and meanwhile yield even better performance. However, there are still several methods for which our EasyRobust cannot get equivalent effect. For instance, our results have gaps with the reported performance of PRIME, Debiased and AdvProp yet. But these gaps seem to be acceptable, and we will make more efforts to fix the gap in the later optimized version of EasyRobust. On ImageNet-C, the best result of EasyRobust is from the DAT method, which conducts discrete adversarial training in the ViT's token space, and achieves 45.59 mCE. It should be denoted that the EasyRobust implementation of DAT has got the public state-of-the-art\footnote{https://paperswithcode.com/sota/domain-generalization-on-imagenet-c}. On Stylized-ImageNet, SIN achieves 59.37\% top@1 accuracy and shows significant improvement. Such a great promotion is attributed to the training augmentation of style transferred images, which is overlapping with the Stylized-ImageNet images. For other OOD benchmarks, WiSE reports the best results. The strong pre-trained CLIP model brings powerful OOD generalization to the downstream fine-tuned WiSE-FT model. Without the help of additional data or pre-trained model, RVT-S$^{*}$ exhibits the most strongest generalization ability, which has exceeded all ResNet50 models with a large margin.

\section{Discussion}

\subsection{Limitations}
So far, our EasyRobust only supports the robust methods on image classification task. As a basic visual task, image classification has a simple pipline and is suitable for academic research of robustness. However, general computer vision in practice also contains more other tasks, such as object detection, semantic and instance segmentation, image retrieval and etc. The robust vision on these tasks is still important but has not been thoroughly studied. On the other hand, visual tasks with stronger application value will have larger significance for robustness construction. 

\subsection{Future Directions} First expansion direction of EasyRobust is the support of more SOTA methods. There are some papers reporting more advanced performance in new year of ICLR2023, CVPR2023, ICML2023 and etc. It is necessary for EasyRobust to follow the latest research. For supported methods in EasyRobust, the effectiveness can also be further improved. Another direction is the comprehensive benchmarks provided for other more visual tasks. Our future development of EasyRobust will support robustness benchmarks on various visual tasks, to accelerate further robustness research in these areas. 

\section{Conclusion}
We introduce EasyRobust, an open source library for industrial application and academic research on robust and generalized vision. EasyRobust builds comprehensive benchmarks and implements most SOTA methods on both adversarial robustness and non-adversarial robustness. Further more, useful analytical tools are supported for users visualizing and explain their robust models. The experiment suggests that our EasyRobust exhibits superior performance on most robust methods proposed in recent years. EasyRobust version v0.2.0 is released as open-source code under Apache License 2.0. We will provide continuous support for users and work for further improvements in the future versions. We wish our EasyRobust can help for training practically-robust models and promote academic and industrial progress in closing the gap between human and machine vision.


\begin{thebibliography}{00}
\bibitem{b1} He, Kaiming, X. Zhang, Shaoqing Ren and Jian Sun, “Deep Residual Learning for Image Recognition.” 2016 IEEE Conference on Computer Vision and Pattern Recognition (CVPR) (2015): 770-778.
\bibitem{b2} Girshick, Ross B., “Fast R-CNN.” 2015 IEEE International Conference on Computer Vision (ICCV) (2015): 1440-1448.
\bibitem{b3} Ren, Shaoqing, Kaiming He, Ross B. Girshick and Jian Sun, “Faster R-CNN: Towards Real-Time Object Detection with Region Proposal Networks.” IEEE Transactions on Pattern Analysis and Machine Intelligence 39 (2015): 1137-1149.
\bibitem{b4} Krizhevsky, Alex, Ilya Sutskever and Geoffrey E. Hinton, “ImageNet classification with deep convolutional neural networks.” Communications of the ACM 60 (2012): 84 - 90.
\bibitem{b5} Szegedy, Christian, Wojciech Zaremba, Ilya Sutskever, Joan Bruna, D. Erhan, Ian J. Goodfellow and Rob Fergus, “Intriguing properties of neural networks.” CoRR abs/1312.6199 (2013): n. pag.
\bibitem{b6} Taori, Rohan, Achal Dave, Vaishaal Shankar, Nicholas Carlini, Benjamin Recht and Ludwig Schmidt, "Measuring robustness to natural distribution shifts in image classification." Advances in Neural Information Processing Systems 33 (2020): 18583-18599.
\bibitem{b7} Hendrycks, Dan, and Thomas Dietterich, "Benchmarking Neural Network Robustness to Common Corruptions and Perturbations." International Conference on Learning Representations (2019).
\bibitem{b8} Croce, Francesco, and Matthias Hein, "Reliable evaluation of adversarial robustness with an ensemble of diverse parameter-free attacks." International conference on machine learning. PMLR, 2020.
\bibitem{b9} Hendrycks, Dan, Kevin Zhao, Steven Basart, Jacob Steinhardt and Dawn Xiaodong Song, “Natural Adversarial Examples.” 2021 IEEE/CVF Conference on Computer Vision and Pattern Recognition (CVPR) (2019): 15257-15266.
\bibitem{b10} Hendrycks, Dan, Steven Basart, Norman Mu, Saurav Kadavath, Frank Wang, Evan Dorundo, Rahul Desai, Tyler Lixuan Zhu, Samyak Parajuli, Mike Guo, Dawn Xiaodong Song, Jacob Steinhardt and Justin Gilmer, “The Many Faces of Robustness: A Critical Analysis of Out-of-Distribution Generalization.” 2021 IEEE/CVF International Conference on Computer Vision (ICCV) (2020): 8320-8329.
\bibitem{b11} Wang, Haohan, Songwei Ge, Eric P. Xing and Zachary Chase Lipton, “Learning Robust Global Representations by Penalizing Local Predictive Power.” Neural Information Processing Systems (2019).
\bibitem{b12} Barbu, Andrei, David Mayo, Julian Alverio, William Luo, Christopher Wang, Dan Gutfreund, Joshua B. Tenenbaum and Boris Katz, “ObjectNet: A large-scale bias-controlled dataset for pushing the limits of object recognition models.” Neural Information Processing Systems (2019).
\bibitem{b13} Recht, Benjamin, Rebecca Roelofs, Ludwig Schmidt and Vaishaal Shankar, “Do ImageNet Classifiers Generalize to ImageNet?” International Conference on Machine Learning (2019).
\bibitem{b14} Madry, Aleksander, Aleksandar Makelov, Ludwig Schmidt, Dimitris Tsipras and Adrian Vladu, "Towards Deep Learning Models Resistant to Adversarial Attacks." International Conference on Learning Representations (2017).
\bibitem{b15} Nie, Weili, Brandon Guo, Yujia Huang, Chaowei Xiao, Arash Vahdat and Anima Anandkumar, "Diffusion Models for Adversarial Purification." International Conference on Machine Learning. PMLR, 2022.
\bibitem{b16} Zhang, Hongyang, Yaodong Yu, Jiantao Jiao, Eric P. Xing, Laurent El Ghaoui and Michael I. Jordan, "Theoretically principled trade-off between robustness and accuracy." International conference on machine learning. PMLR, 2019.
\bibitem{b17} Guo, Minghao, Yuzhe Yang, Rui Xu and Ziwei Liu, “When NAS Meets Robustness: In Search of Robust Architectures Against Adversarial Attacks.” 2020 IEEE/CVF Conference on Computer Vision and Pattern Recognition (CVPR) (2019): 628-637.
\bibitem{b18} Zhang, Hongyi, Moustapha Cissé, Yann Dauphin and David Lopez-Paz, "mixup: Beyond Empirical Risk Minimization." International Conference on Learning Representations (2017).
\bibitem{b19} Geirhos, Robert, Patricia Rubisch, Claudio Michaelis, Matthias Bethge, Felix Wichmann and Wieland Brendel, "ImageNet-trained CNNs are biased towards texture; increasing shape bias improves accuracy and robustness." International Conference on Learning Representations (2018).
\bibitem{b20} Xie, Cihang, Mingxing Tan, Boqing Gong, Jiang Wang, Alan Loddon Yuille and Quoc V. Le, “Adversarial Examples Improve Image Recognition.” 2020 IEEE/CVF Conference on Computer Vision and Pattern Recognition (CVPR) (2019): 816-825.
\bibitem{b21} Herrmann, Charles, Kyle Sargent, Lu Jiang, Ramin Zabih, Huiwen Chang, Ce Liu, Dilip Krishnan and Deqing Sun, “Pyramid Adversarial Training Improves ViT Performance.” 2022 IEEE/CVF Conference on Computer Vision and Pattern Recognition (CVPR) (2021): 13409-13419.
\bibitem{b22} Mao, Xiaofeng, Yuefeng Chen, Ranjie Duan, Yao Zhu, Gege Qi, Shaokai Ye, Xiaodan Li, Rong Zhang and Hui Xue, “Enhance the Visual Representation via Discrete Adversarial Training.” Neural Information Processing Systems (2022).
\bibitem{b23} Goodfellow, Ian J., Nicolas Papernot and Patrick Mcdaniel, “Cleverhans V0.1: an Adversarial Machine Learning Library.” unpublished.
\bibitem{b24} Rauber, Jonas, Wieland Brendel and Matthias Bethge, “Foolbox: A Python toolbox to benchmark the robustness of machine learning models.” unpublished.
\bibitem{b25} Nicolae, Maria-Irina, Mathieu Sinn, Minh-Ngoc Tran, Beat Buesser, Ambrish Rawat, Martin Wistuba, Valentina Zantedeschi, Nathalie Baracaldo, Bryant Chen, Heiko Ludwig, Ian Molloy and Ben Edwards, “Adversarial Robustness Toolbox v1.0.0.” unpublished.
\bibitem{b26} Dong, Yinpeng, Qi-An Fu, Xiao Yang, Tianyu Pang, Hang Su, Zihao Xiao and Jun Zhu, “Benchmarking Adversarial Robustness on Image Classification.” 2020 IEEE/CVF Conference on Computer Vision and Pattern Recognition (CVPR) (2020): 318-328.
\bibitem{b27} Tang, Shiyu, Ruihao Gong, Yan Wang, Aishan Liu, Jiakai Wang, Xinyun Chen, Fengwei Yu, Xianglong Liu, Dawn Xiaodong Song, Alan Loddon Yuille, Philip H. S. Torr and Dacheng Tao, “RobustART: Benchmarking Robustness on Architecture Design and Training Techniques.” in press.
\bibitem{b28} Gulrajani, Ishaan, and David Lopez-Paz, "In Search of Lost Domain Generalization." International Conference on Learning Representations (2022).
\bibitem{b29} Croce, Francesco, Maksym Andriushchenko, Vikash Sehwag, Edoardo Debenedetti, Edoardo Debenedetti, Mung Chiang, Prateek Mittal and Matthias Hein, "RobustBench: a standardized adversarial robustness benchmark." Thirty-fifth Conference on Neural Information Processing Systems Datasets and Benchmarks Track (Round 2) (2020).
\bibitem{b30} Geirhos, Robert, Kantharaju Narayanappa, Benjamin Mitzkus, Tizian Thieringer, Matthias Bethge, Felix A. Wichmann and Wieland Brendel, “Partial success in closing the gap between human and machine vision.” Neural Information Processing Systems (2021).
\bibitem{b31} Liu, Ze, Yutong Lin, Yue Cao, Han Hu, Yixuan Wei, Zheng Zhang, Stephen Lin and Baining Guo, “Swin Transformer: Hierarchical Vision Transformer using Shifted Windows.” 2021 IEEE/CVF International Conference on Computer Vision (ICCV) (2021): 9992-10002.
\bibitem{b32} He, Kaiming, Xinlei Chen, Saining Xie, Yanghao Li, Piotr Doll'ar and Ross B. Girshick, “Masked Autoencoders Are Scalable Vision Learners.” 2022 IEEE/CVF Conference on Computer Vision and Pattern Recognition (CVPR) (2021): 15979-15988.
\bibitem{b33} Andriushchenko, Maksym, Francesco Croce, Nicolas Flammarion and Matthias Hein, "Square attack: a query-efficient black-box adversarial attack via random search." Computer Vision–ECCV 2020: 16th European Conference, Glasgow, UK, August 23–28, 2020, Proceedings, Part XXIII. Cham: Springer International Publishing, 2020.
\bibitem{b34} Croce, Francesco and Matthias Hein, “Minimally distorted Adversarial Examples with a Fast Adaptive Boundary Attack.” International Conference on Machine Learning (2019).
\bibitem{b35} Hendrycks, Dan, Norman Mu, Ekin Dogus Cubuk, Barret Zoph, Justin Gilmer and Balaji Lakshminarayanan, “AugMix: A Simple Data Processing Method to Improve Robustness and Uncertainty.” International Conference on Learning Representations (2019).
\bibitem{b36} Chen, Guangyao, Peixi Peng, Li Ma, Jia Li, Lin Du and Yonghong Tian, “Amplitude-Phase Recombination: Rethinking Robustness of Convolutional Neural Networks in Frequency Domain.” 2021 IEEE/CVF International Conference on Computer Vision (ICCV) (2021): 448-457.
\bibitem{b37} Foret, Pierre, Ariel Kleiner, Hossein Mobahi and Behnam Neyshabur, “Sharpness-Aware Minimization for Efficiently Improving Generalization.” International Conference on Learning Representations (2020).
\bibitem{b38} Zhuang, Juntang, Boqing Gong, Liangzhe Yuan, Yin Cui, Hartwig Adam, Nicha C. Dvornek, Sekhar Chandra Tatikonda, James Duncan and Ting Liu, “Surrogate Gap Minimization Improves Sharpness-Aware Training.” International Conference on Learning Representations (2022).
\bibitem{b39} Du, Jiawei, Hanshu Yan, Jiashi Feng, Joey Tianyi Zhou, Liangli Zhen, Rick Siow Mong Goh and Vincent Y. F. Tan, “Efficient Sharpness-aware Minimization for Improved Training of Neural Networks.” International Conference on Learning Representations (2021).
\bibitem{b40} Dosovitskiy, Alexey, Lucas Beyer, Alexander Kolesnikov, Dirk Weissenborn, Xiaohua Zhai, Thomas Unterthiner, Mostafa Dehghani, Matthias Minderer, Georg Heigold, Sylvain Gelly, Jakob Uszkoreit and Neil Houlsby, “An Image is Worth 16x16 Words: Transformers for Image Recognition at Scale.” International Conference on Learning Representations (2020).
\bibitem{b41} Li, Qiufu, Linlin Shen, Sheng Guo and Zhihui Lai, “WaveCNet: Wavelet Integrated CNNs to Suppress Aliasing Effect for Noise-Robust Image Classification.” IEEE Transactions on Image Processing 30 (2021): 7074-7089.
\bibitem{b42} Mao, Xiaofeng, Gege Qi, Yuefeng Chen, Xiaodan Li, Ranjie Duan, Shaokai Ye, Yuan He and Hui Xue, “Towards Robust Vision Transformer.” 2022 IEEE/CVF Conference on Computer Vision and Pattern Recognition (CVPR) (2021): 12032-12041.
\bibitem{b43} Mao, Chengzhi, Lu Jiang, Mostafa Dehghani, Carl Vondrick, Rahul Sukthankar and Irfan Essa. “Discrete Representations Strengthen Vision Transformer Robustness.” International Conference on Learning Representations (2021).
\bibitem{b44} Zhou, Daquan, Zhiding Yu, Enze Xie, Chaowei Xiao, Anima Anandkumar, Jiashi Feng and José Manuel Álvarez. “Understanding The Robustness in Vision Transformers.” International Conference on Machine Learning (2022)
\bibitem{b45} Nuriel, Oren, Sagie Benaim and Lior Wolf, “Permuted AdaIN: Reducing the Bias Towards Global Statistics in Image Classification.” 2021 IEEE/CVF Conference on Computer Vision and Pattern Recognition (CVPR) (2020): 9477-9486.
\bibitem{b46} Tang, Zhiqiang, Yunhe Gao, Yi Zhu, Zhi Zhang, Mu Li and Dimitris N. Metaxas, “CrossNorm and SelfNorm for Generalization under Distribution Shifts.” 2021 IEEE/CVF International Conference on Computer Vision (ICCV) (2021): 52-61.
\bibitem{b47} Radford, Alec, Jong Wook Kim, Chris Hallacy, Aditya Ramesh, Gabriel Goh, Sandhini Agarwal, Girish Sastry, Amanda Askell, Pamela Mishkin, Jack Clark, Gretchen Krueger and Ilya Sutskever, “Learning Transferable Visual Models From Natural Language Supervision.” International Conference on Machine Learning (2021).
\bibitem{b48} Wortsman, Mitchell, Gabriel Ilharco, Mike Li, Jong Wook Kim, Hannaneh Hajishirzi, Ali Farhadi, Hongseok Namkoong and Ludwig Schmidt, “Robust fine-tuning of zero-shot models.” 2022 IEEE/CVF Conference on Computer Vision and Pattern Recognition (CVPR) (2021): 7949-7961.
\bibitem{b49} Chen, Sizhe, Zhengbao He, Chengjin Sun and Xiaolin Huang, “Universal Adversarial Attack on Attention and the Resulting Dataset DAmageNet.” IEEE Transactions on Pattern Analysis and Machine Intelligence 44 (2020): 2188-2197.
\bibitem{b50} Pintor, Maura, Daniele Angioni, Angelo Sotgiu, Luca Demetrio, Ambra Demontis, Battista Biggio and Fabio Roli, “ImageNet-Patch: A Dataset for Benchmarking Machine Learning Robustness against Adversarial Patches.” Pattern Recognit. 134 (2022): 109064.
\bibitem{b51} Zhou, Bolei, Aditya Khosla, Àgata Lapedriza, Aude Oliva and Antonio Torralba, “Learning Deep Features for Discriminative Localization.” 2016 IEEE Conference on Computer Vision and Pattern Recognition (CVPR) (2015): 2921-2929.
\bibitem{b52} Selvaraju, Ramprasaath R., Abhishek Das, Ramakrishna Vedantam, Michael Cogswell, Devi Parikh and Dhruv Batra, “Grad-CAM: Visual Explanations from Deep Networks via Gradient-Based Localization.” International Journal of Computer Vision 128 (2016): 336-359.
\bibitem{b53} Chattopadhyay, Aditya, Anirban Sarkar, Prantik Howlader and Vineeth N. Balasubramanian, “Grad-CAM++: Generalized Gradient-Based Visual Explanations for Deep Convolutional Networks.” 2018 IEEE Winter Conference on Applications of Computer Vision (WACV) (2017): 839-847.
\bibitem{b54} Fu, Ruigang, Qingyong Hu, Xiaohu Dong, Yulan Guo, Yinghui Gao and Biao Li, “Axiom-based Grad-CAM: Towards Accurate Visualization and Explanation of CNNs.” British Machine Vision Conference (2020).
\bibitem{b55} Draelos, Rachel Lea and Lawrence Carin. “HiResCAM: Explainable Multi-Organ Multi-Abnormality Prediction in 3D Medical Images.” unpublished.
\bibitem{b56} Wang, Haofan, Zifan Wang, Mengnan Du, Fan Yang, Zijian Zhang, Sirui Ding, Piotr (Peter) Mardziel and Xia Hu. “Score-CAM: Score-Weighted Visual Explanations for Convolutional Neural Networks.” 2020 IEEE/CVF Conference on Computer Vision and Pattern Recognition Workshops (CVPRW) (2019): 111-119.
\bibitem{b57} Desai, Saurabh Satish and H. G. Ramaswamy. “Ablation-CAM: Visual Explanations for Deep Convolutional Network via Gradient-free Localization.” 2020 IEEE Winter Conference on Applications of Computer Vision (WACV) (2020): 972-980.
\bibitem{b58} Muhammad, Mohammed Bany and Mohammed Yeasin. “Eigen-CAM: Class Activation Map using Principal Components.” 2020 International Joint Conference on Neural Networks (IJCNN) (2020): 1-7.
\bibitem{b59} Jiang, Peng-Tao, Chang-Bin Zhang, Qibin Hou, Ming-Ming Cheng and Yunchao Wei. “LayerCAM: Exploring Hierarchical Class Activation Maps for Localization.” IEEE Transactions on Image Processing 30 (2021): 5875-5888.
\bibitem{b60} Srinivas, Suraj and François Fleuret. “Full-Gradient Representation for Neural Network Visualization.” Neural Information Processing Systems (2019).
\bibitem{b61} Abnar, Samira and Willem Zuidema. “Quantifying Attention Flow in Transformers.” Annual Meeting of the Association for Computational Linguistics (2020).
\bibitem{b62} Cubuk, Ekin Dogus, Barret Zoph, Jonathon Shlens and Quoc V. Le. “Randaugment: Practical automated data augmentation with a reduced search space.” 2020 IEEE/CVF Conference on Computer Vision and Pattern Recognition Workshops (CVPRW) (2019): 3008-3017.
\bibitem{b63} Yun, Sangdoo, Dongyoon Han, Seong Joon Oh, Sanghyuk Chun, Junsuk Choe and Young Joon Yoo. “CutMix: Regularization Strategy to Train Strong Classifiers With Localizable Features.” 2019 IEEE/CVF International Conference on Computer Vision (ICCV) (2019): 6022-6031.
\bibitem{b64} Wong, Eric, Leslie Rice and J. Zico Kolter. “Fast is better than free: Revisiting adversarial training.” International Conference on Learning Representations (2020).
\bibitem{b65} Shafahi, Ali, Mahyar Najibi, Amin Ghiasi, Zheng Xu, John P. Dickerson, Christoph Studer, Larry S. Davis, Gavin Taylor and Tom Goldstein. “Adversarial Training for Free!” Neural Information Processing Systems (2019).
\bibitem{b66} Debenedetti, Edoardo, Vikash Sehwag and Prateek Mittal. “A Light Recipe to Train Robust Vision Transformers.” First IEEE Conference on Secure and Trustworthy Machine Learning (2022).
\bibitem{b67} Bai, Yutong, Jieru Mei, Alan Loddon Yuille and Cihang Xie. “Are Transformers More Robust Than CNNs?” Neural Information Processing Systems (2021).
\bibitem{b68} El-Nouby, Alaaeldin, Hugo Touvron, Mathilde Caron, Piotr Bojanowski, Matthijs Douze, Armand Joulin, Ivan Laptev, Natalia Neverova, Gabriel Synnaeve, Jakob Verbeek and Hervé Jégou. “XCiT: Cross-Covariance Image Transformers.” Neural Information Processing Systems (2021).
\bibitem{b69} Hu, Jie, Li Shen, Samuel Albanie, Gang Sun and Enhua Wu. “Squeeze-and-Excitation Networks.” IEEE Transactions on Pattern Analysis and Machine Intelligence 42 (2017): 2011-2023.
\bibitem{b70} Tan, Mingxing and Quoc V. Le. “EfficientNet: Rethinking Model Scaling for Convolutional Neural Networks.” International Conference on Machine Learning (2019).
\bibitem{b71} Simonyan, Karen and Andrew Zisserman. “Very Deep Convolutional Networks for Large-Scale Image Recognition.” International Conference on Learning Representations (2014).
\bibitem{b72} Xie, Cihang, Mingxing Tan, Boqing Gong, Alan Loddon Yuille and Quoc V. Le. “Smooth Adversarial Training.” unpublished.
\bibitem{b73} Wang, Haotao, Aston Zhang, Shuai Zheng, Xingjian Shi, Mu Li and Zhangyang Wang. “Removing Batch Normalization Boosts Adversarial Training.” International Conference on Machine Learning (2022).
\bibitem{b74} Modas, Apostolos, Rahul Rade, Guillermo Ortiz-Jiménez, Seyed-Mohsen Moosavi-Dezfooli and Pascal Frossard. “PRIME: A Few Primitives Can Boost Robustness to Common Corruptions.” European Conference on Computer Vision (2021).
\bibitem{b75} Li, Yingwei, Qihang Yu, Mingxing Tan, Jieru Mei, Peng Tang, Wei Shen, Alan Loddon Yuille and Cihang Xie. “Shape-Texture Debiased Neural Network Training.” International Conference on Learning Representations (2020).
\bibitem{b76} Zhang, Richard. “Making Convolutional Networks Shift-Invariant Again.” International Conference on Machine Learning (2019).
\bibitem{b77} Lim, Bee, Sanghyun Son, Heewon Kim, Seungjun Nah and Kyoung Mu Lee. “Enhanced Deep Residual Networks for Single Image Super-Resolution.” 2017 IEEE Conference on Computer Vision and Pattern Recognition Workshops (CVPRW) (2017): 1132-1140.
\bibitem{b78} Theis, Lucas, Wenzhe Shi, Andrew Cunningham and Ferenc Huszár. “Lossy Image Compression with Compressive Autoencoders.” International Conference on Learning Representations (2017).
\bibitem{b79} Xie, Cihang and Alan Loddon Yuille. “Intriguing Properties of Adversarial Training at Scale.” International Conference on Learning Representations (2019).
\end{thebibliography}
\end{document}